%% file: acl_latex.tex
\pdfoutput=1
\documentclass[11pt]{article}

\usepackage[final]{acl}

\usepackage{times}
\usepackage{latexsym}
\usepackage{xspace}
\usepackage{booktabs}

\usepackage{float}

\usepackage{amsmath}

\usepackage{listings}

\usepackage[T1]{fontenc}
\usepackage[utf8]{inputenc}

\usepackage{microtype}

\usepackage{inconsolata}

\usepackage{graphicx}

\usepackage{multirow}
\usepackage{hhline}
\usepackage{cleveref}

\usepackage[normalem]{ulem}

\input{macro}

\title{EquiBench: Benchmarking Large Language Models' Reasoning about Program Semantics via Equivalence Checking}

\author{
\\
\textbf{Anjiang Wei}\textsuperscript{1}\thanks{Correspondence to: anjiang@cs.stanford.edu} \hspace{1em} \textbf{Jiannan Cao}\textsuperscript{2} \hspace{1em} \textbf{Ran Li}\textsuperscript{1,3} \hspace{1em} \textbf{Hongyu Chen}\textsuperscript{4} \hspace{1em} \textbf{Yuhui Zhang}\textsuperscript{1} \\
\textbf{Ziheng Wang}\textsuperscript{1} \hspace{1em} \textbf{Yuan Liu}\textsuperscript{3} \hspace{1em} \textbf{Thiago S. F. X. Teixeira}\textsuperscript{5} \\
\textbf{Diyi Yang}\textsuperscript{1} \hspace{1em} \textbf{Ke Wang}\textsuperscript{4} \hspace{1em} \textbf{Alex Aiken}\textsuperscript{1} \\
\\
\textsuperscript{1}Stanford University \hspace{1em} \textsuperscript{2}MIT \hspace{1em} \textsuperscript{3}Google \hspace{1em} \textsuperscript{4}Nanjing University \hspace{1em} \textsuperscript{5}Intel \hspace{1em}
}

\begin{document}
\maketitle
\begin{abstract}

% involved macros:
% \name, \numpair, \numllm, \sotaacc{}, \sotacuda{}, \sotadce{}

As large language models (LLMs) become integral to code-related tasks, a central question emerges: Do LLMs truly understand program semantics? We introduce EquiBench, a new benchmark for evaluating LLMs through equivalence checking, i.e., determining whether two programs produce identical outputs for all possible inputs. Unlike prior code generation benchmarks, this task directly tests a model’s ability to reason about program semantics. EquiBench consists of 2400 program pairs across four languages and six categories. These pairs are generated through program analysis, compiler scheduling, and superoptimization, ensuring high-confidence labels, nontrivial difficulty, and full automation. We evaluate 19 state-of-the-art LLMs and find that in the most challenging categories, the best accuracies are 63.8\% and 76.2\%, only modestly above the 50\% random baseline. Further analysis reveals that models often rely on syntactic similarity rather than exhibiting robust reasoning about program semantics, highlighting current limitations. Our code and dataset are publicly available at \url{https://github.com/Anjiang-Wei/equibench}

\end{abstract}

\input{1_intro}

\input{2_related}

\input{3_method}

\input{4_setup}

\input{5_result}

\input{6_discussion}

\input{7_conclusion}

\section*{Limitations}
While we make every effort to ensure that all program pairs in \name{} are correctly labeled, we cannot guarantee absolute accuracy. The dataset is built through automated transformation pipelines that rely on external toolchains such as compilers, superoptimizers, and analysis frameworks. These components, although carefully chosen for their soundness guarantees, are not immune to subtle bugs or rare edge cases that may produce incorrect outputs. Furthermore, in the categories based on competitive programming submissions, some input programs may themselves be mislabeled due to incorrect online judge verdicts or subtle implementation flaws that escape the test cases. These sources of noise, though limited in scope, remain a potential source of labeling inaccuracy.

\section*{Acknowledgments}
We thank Mingfei Guo, Lianmin Zheng, Allen Nie, Shiv Sundram, and Xiaohan Wang for their discussions. This work was partially supported by a Google Research Award and OpenAI’s Researcher Access Program.

\bibliography{custom}

\setcounter{figure}{0}
\renewcommand{\thefigure}{A\arabic{figure}}
\setcounter{table}{0}
\renewcommand{\thetable}{A\arabic{table}}

\newpage
\appendix
% \onecolumn

\input{8_appendix}

\end{document}

%% file: macro.tex
\newcommand{\CodeIn}[1]{\texttt{#1}}

\newcommand{\numpair}{2400\xspace}

\newcommand{\name}{EquiBench\xspace}

\newcommand{\sotaacc}{82.3\%\xspace}
\newcommand{\sotalowacc}{60.8\%\xspace}

\newcommand{\sotadce}{76.2\%\xspace}
\newcommand{\sotacuda}{63.8\%\xspace}
\newcommand{\sotaojv}{96.5\%\xspace}

\newcommand{\dce}{DCE\xspace}
\newcommand{\cuda}{CUDA\xspace}
\newcommand{\ass}{x86-64\xspace}
\newcommand{\oja}{OJ\_A\xspace}
\newcommand{\ojv}{OJ\_V\xspace}
\newcommand{\ojva}{OJ\_VA\xspace}

\newcommand{\numllm}{19\xspace}

\lstdefinestyle{mystyle}{
    language=Python,
    basicstyle=\ttfamily\small,
     morekeywords={char,static,true,false},
    numbersep=4pt,
    commentstyle=\color{gray}
}

%% file: 1_intro.tex
\section{Introduction}
\label{sec:intro}

\begin{figure*}[!tb]
\centering
\includegraphics[width=\textwidth]{figure/overview.pdf}
\caption{\textbf{Overview of \name.} We construct (in)equivalent program pairs from diverse sources, including C and CUDA programs, x86-64 assembly, and competitive programming, using automated transformations based on program analysis, compiler scheduling, superoptimization, and changes in algorithms or variable names.}
\label{fig:overview}
\end{figure*}

Large language models (LLMs) have rapidly become central to software engineering workflows, powering tools for code generation, program repair, test case generation, debugging, and beyond, significantly boosting developers' productivity~\cite{jain2024livecodebench,yang2024whitefox,yang2023kernelgpt}. This surge of capability has prompted a natural yet fundamental question: Do LLMs merely mimic code syntax they have seen during training, or do they genuinely understand what programs do?

Unlike natural language, code is executable. Two programs may differ syntactically yet be semantically equivalent, producing identical outputs for all inputs. Conversely, programs with only minor syntactic differences can behave quite differently at runtime. This gap between surface-level program features and actual execution behavior raises an important question: Does training on static code corpora equip LLMs with a grounded understanding of program semantics?

To rigorously assess whether LLMs truly understand code, we need benchmarks that demand reasoning about program semantics. However, widely used coding benchmarks such as HumanEval~\cite{chen2021evaluating} and MBPP~\cite{austin2021program} primarily test a model’s ability to generate short code snippets from natural language descriptions, offering limited insight into whether the model grasps the underlying semantics of the code it generates.

In this work, we introduce \textbf{equivalence checking} as a new task for evaluating LLMs' ability to reason about program semantics. Unlike tasks based on syntactic similarity, equivalence checking asks whether two programs are semantically equivalent, i.e., whether they produce identical outputs for all possible inputs, regardless of how differently they are written. Program equivalence problems test directly whether and how well models reason about code. Any question about
program semantics can be formulated as an equivalence checking problem, and program equivalence
problems can have any level of difficulty from trivially easy to extremely difficult. Program equivalence is undecidable in general: no algorithm can determine program equivalence for all cases while guaranteeing termination. This fundamental theoretical impossibility underscores the intrinsic difficulty of our task.

Designing a benchmark for equivalence checking requires both equivalent and inequivalent program pairs spanning diverse categories, which poses several challenges in terms of \emph{label soundness}, \emph{problem difficulty}, and \emph{automation}. First, it is difficult to guarantee high-confidence labels, as verifying equivalence by exhaustively executing all possible inputs is almost always computationally infeasible. Second, existing generation techniques rely on superficial syntactic edits~\cite{badihi2021eqbench,maveli2024can}, which are too simplistic to meaningfully challenge state-of-the-art LLMs and fail to probe their semantic reasoning limits. Third, to enable comprehensive evaluation, the benchmark must be large-scale and modular, necessitating a fully automated construction pipeline.

In this work, we introduce \textbf{\name{}}, a dataset of \numpair program pairs for evaluating large language models on equivalence checking. Covering Python, C, CUDA, and x86-64 programs, it enables a systematic assessment of LLMs' ability to reason about program semantics.

As illustrated in \Cref{fig:overview}, \name{} addresses these challenges by automatically constructing both equivalent and inequivalent program pairs from diverse input sources, including randomly generated C and CUDA code, assembly instructions, and competitive programming solutions. To ensure label soundness without exhaustive execution, we apply program transformation techniques grounded in program analysis and superoptimization. To increase problem difficulty beyond trivial edits, we incorporate structural transformations through compiler scheduling and algorithmic equivalences. The entire generation pipeline is fully automated, enabling scalable construction of a large and diverse benchmark. Finally, \name{} is extensible to additional categories of equivalence checking problems, which we anticipate will be useful as LLMs improve.

Our experiments show that \name{} is a challenging benchmark for LLMs. Among the \numllm models evaluated, OpenAI o4-mini performs best overall, yet achieves only \sotalowacc{} in the CUDA category despite reaching the highest overall accuracy of \sotaacc{}. In the two most difficult categories, the best accuracies are \sotacuda{} and \sotadce{}, respectively, only modestly better than the random baseline of 50\% for binary classification. In contrast, purely syntactic changes such as variable renaming are much easier, with accuracies as high as \sotaojv{}. We further find, through difficulty analysis, that models often rely on superficial form features such as syntactic similarity rather than demonstrating robust semantic reasoning. Moreover, prompting strategies such as few-shot in-context learning and Chain-of-Thought (CoT) prompting barely improve LLM performance, underscoring the difficulty of reasoning about program semantics.

In summary, our contributions are as follows:

\begin{itemize}
    \item \textbf{New Task and Dataset:} We introduce equivalence checking as a new task to assess LLMs' reasoning about program semantics. We present \textit{\name{}}, a benchmark for equivalence checking spanning four languages and six equivalence categories.
    
    \item \textbf{Automated Generation:} We develop a fully automated pipeline for constructing diverse (in)equivalent program pairs using techniques that ensure high-confidence labels and nontrivial difficulty. The pipeline covers transformations ranging from syntactic edits to structural modifications and algorithmic equivalence.

    \item \textbf{Evaluation and Analysis:} We evaluate \numllm state-of-the-art models on \name. In the two most challenging categories, the best accuracies are only \sotacuda{} and \sotadce{}, highlighting fundamental limitations. Our analysis shows that models often rely on superficial form features rather than demonstrating robust reasoning about program semantics.
\end{itemize}

%% file: 2_related.tex
\section{Related Work}
\label{sec:related}

\paragraph{LLM Reasoning Benchmarks} Extensive research has evaluated LLMs' reasoning capabilities across diverse tasks~\cite{cobbe2021training,huang2022towards,bubeck2023sparks,mirzadeh2024gsm,zhou2022least,ho2022large,wei2022chain,chen2024large,clark2018think,zhang2024transformer}. In the context of code reasoning, i.e., predicting a program's execution behavior without running it, CRUXEval~\cite{gu2024cruxeval} focuses on input-output prediction, while CodeMind~\cite{liu2024codemind} extends evaluation to natural language specifications. Another line of work seeks to improve LLMs' code simulation abilities through prompting~\cite{la2024code} or targeted training~\cite{liu2023code,ni2024next,ding2024semcoder,chen2025dce}. Unlike prior work that tests LLMs on specific inputs, our benchmark evaluates their ability to reason over all inputs.

\paragraph{Equivalence Checking} Equivalence checking underpins applications such as performance optimization~\cite{shypula2023learning,cummins2023large,cummins2024meta}, code transpilation~\cite{lu2021codexglue,yang2024exploring,ibrahimzada2024repository,pan2024lost}, refactoring~\cite{pailoor2024semantic}, and testing~\cite{felsing2014automating,tian2024large}. Due to its undecidable nature, no algorithm can decide program equivalence for all program pairs while always terminating. Existing techniques~\cite{sharma2013data,dahiya2017black,gupta2018effective,mora2018client,churchill2019semantic,badihi2020ardiff} focus on specific domains, such as SQL query equivalence~\cite{zhao2023llm,ding2023proving,singh2024exploring}. EQBENCH~\cite{badihi2021eqbench} and SeqCoBench~\cite{maveli2024can} are the main datasets for equivalence checking, but have limitations. EQBENCH is too small (272 pairs) for LLM evaluation, while SeqCoBench relies only on statement-level syntactic changes (e.g., renaming variables). In contrast, our work introduces a broader set of equivalence categories, creating a more systematic and challenging benchmark.

%% file: 3_method.tex
\section{Benchmark Construction}
\label{sec:method}

While we have so far discussed the standard notion of equivalence, namely that two programs produce the same output on any input, each benchmark category adopts a more precise definition tailored to its domain. All follow the principle of ``producing the same output given the same input,'' but the exact criteria differ. For example, the CUDA category tolerates small discrepancies from floating-point rounding rather than requiring strict bit-level equivalence. These definitions are grounded in real-world use cases and chosen to capture practical notions of equivalence in each setting. For each category, we provide the corresponding definition in the prompt when testing LLM reasoning. We describe how we generate (in)equivalent pairs across the six categories as follows:

\begin{itemize}
    \item \textbf{\dce}: C program pairs generated via the compiler's dead code elimination (DCE) pass (\Cref{subsec:dce}).

    \item \textbf{\cuda}: CUDA program pairs created by applying different scheduling strategies using a tensor compiler (\Cref{subsec:cuda}).
    
    \item \textbf{\ass}: x86-64 assembly program pairs generated by a superoptimizer (\Cref{subsec:x86}).
    
    \item \textbf{\oja}, \textbf{\ojv}, \textbf{\ojva}: Python program pairs from online judge submissions, featuring algorithmic differences (\oja), variable-renaming transformations (\ojv), and combinations of both (\ojva) (\Cref{subsec:oj}).
\end{itemize}

\begin{figure}[!tb]
\centering
\includegraphics[width=\columnwidth]{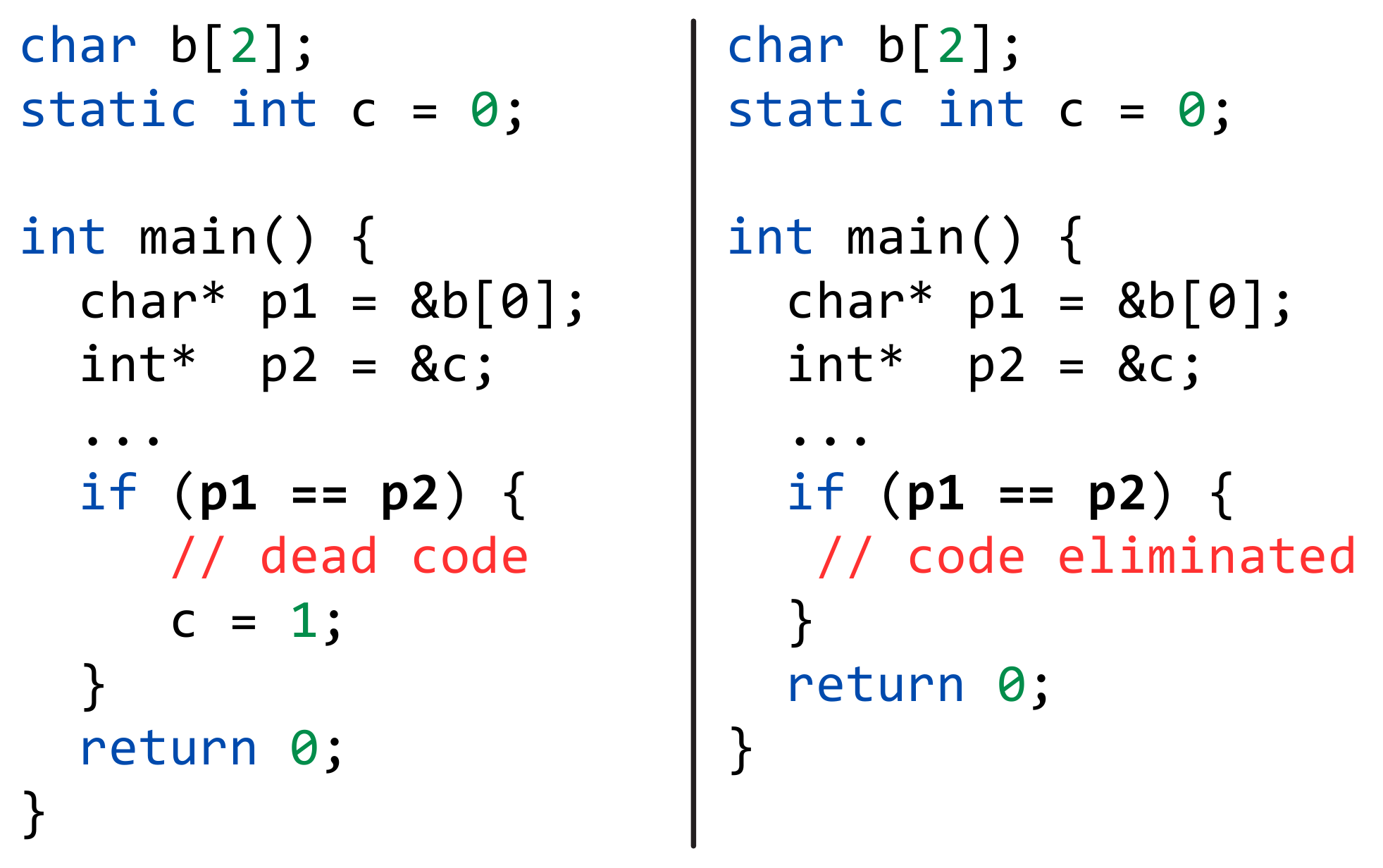}
\caption{\textbf{An equivalent pair from the \dce category in \name.} In the left program, \CodeIn{c = 1} is dead code that has no effect on the program state and is removed in the right program. Such pairs are generated using the Dead Code Elimination (DCE) pass in compilers.}
\label{fig:dce}
\end{figure}

\begin{figure*}[!tb]
\centering
\includegraphics[width=\textwidth]{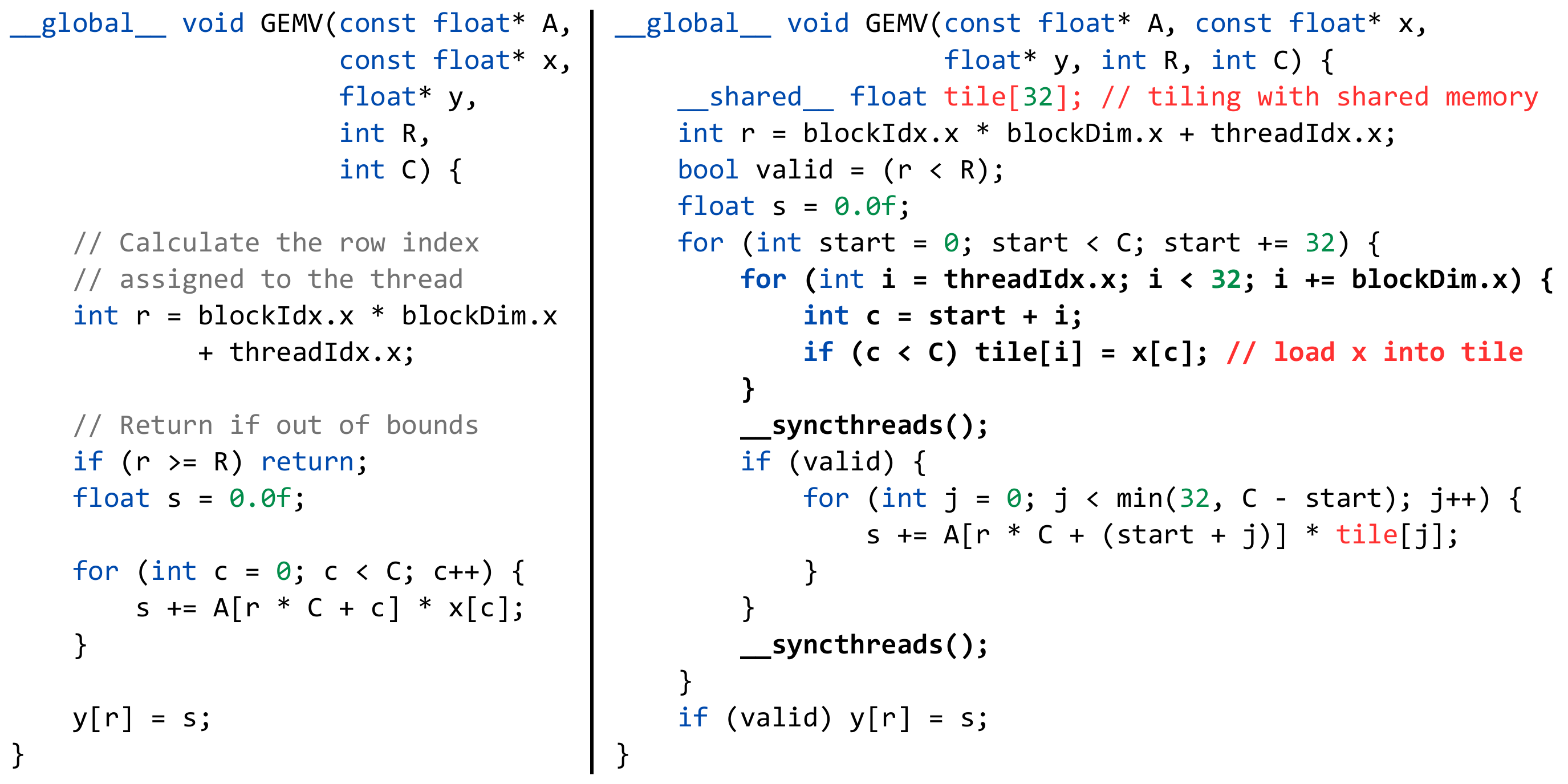}
\caption{\textbf{An equivalent pair from the \cuda category in \name.} Both programs perform matrix-vector multiplication ($y = Ax$). The right-hand program uses \emph{shared memory tiling} to improve performance. Tensor compilers are utilized to explore different \emph{scheduling strategies}, automating the generation.}
\label{fig:cuda}
\end{figure*}

\subsection{Pairs from Program Analysis (DCE)}
\label{subsec:dce}
Dead code elimination (DCE), a compiler pass, removes useless program statements. After DCE, the remaining statements in the modified program naturally {\em correspond} to those in the original program.

\paragraph{Definition of Equivalence.} Two programs are considered equivalent if, when executed on the same input, they \emph{always} have identical \emph{program states} at all corresponding points reachable by program execution. We expect language models to identify differences between the two programs, align their states, and determine whether these states are consistently identical.

\paragraph{Example.} \Cref{fig:dce} illustrates an equivalent pair of C programs. In the left program, the condition (\CodeIn{p1 == p2}) compares the memory address of the first element of the array \CodeIn{b} with that of the static variable \CodeIn{c}. Since \CodeIn{b} and \CodeIn{c} reside in different memory locations, this condition can never be satisfied. As a result, the assignment \CodeIn{c = 1} is never executed in the left program and is removed in the right program.

\paragraph{Automation.} This reasoning process is automated by compilers through \emph{alias analysis}, which statically determines whether two pointers can reference the same memory location. Based on this analysis, the compiler’s \emph{Dead Code Elimination (DCE)} pass removes code that does not affect program semantics to improve performance.

\paragraph{Dataset Generation.} We utilize CSmith~\cite{yang2011finding} to create an initial pool of random C programs. Building on techniques from prior compiler testing research~\cite{theodoridis2022finding}, we implement an LLVM-based tool~\cite{lattner2004llvm} to classify code snippets as either dead or live. Live code is further confirmed by executing random inputs with observable side effects. Equivalent program pairs are generated by eliminating dead code, while inequivalent pairs are generated by removing live code.

\subsection{Pairs from Compiler Scheduling (\cuda)}
\label{subsec:cuda}

\paragraph{Definition of Equivalence.} Two CUDA programs are considered equivalent if they produce the same mathematical output for any valid input, \emph{disregarding floating-point rounding errors}. This definition \emph{differs} from that in \Cref{subsec:dce}, as it does not require the internal program states to be identical during execution.

\paragraph{Example.} \Cref{fig:cuda} shows an equivalent CUDA program pair. Both compute matrix-vector multiplication  $y=Ax$, where $A$ has dimensions (R, C) and $x$ has size C. The right-hand program applies the \emph{shared memory tiling} technique, loading \CodeIn{x} into shared memory \CodeIn{tile} (declared with \CodeIn{\_\_shared\_\_}). Synchronization primitives \CodeIn{\_\_syncthreads()} are properly inserted to prevent synchronization issues.

\paragraph{Automation.} The program transformation can be automated with tensor compilers, which provide a set of \emph{schedules} to optimize loop-based programs. These schedules include loop tiling, loop fusion, loop reordering, loop unrolling, vectorization, and cache optimization. For any given schedule, the compiler can generate the transformed code. While different schedules can significantly impact program performance on the GPU, they do not affect the program's correctness (assuming no compiler bugs), providing the foundation for automation.

\paragraph{Dataset Generation.} We utilize TVM as the tensor compiler~\cite{chen2018tvm} and sample tensor program schedules from TenSet~\cite{zheng2021tenset} to generate equivalent CUDA program pairs. Inequivalent pairs are created by sampling code from different tensor programs.

\begin{figure}[!tb]
\centering
\includegraphics[width=\columnwidth]{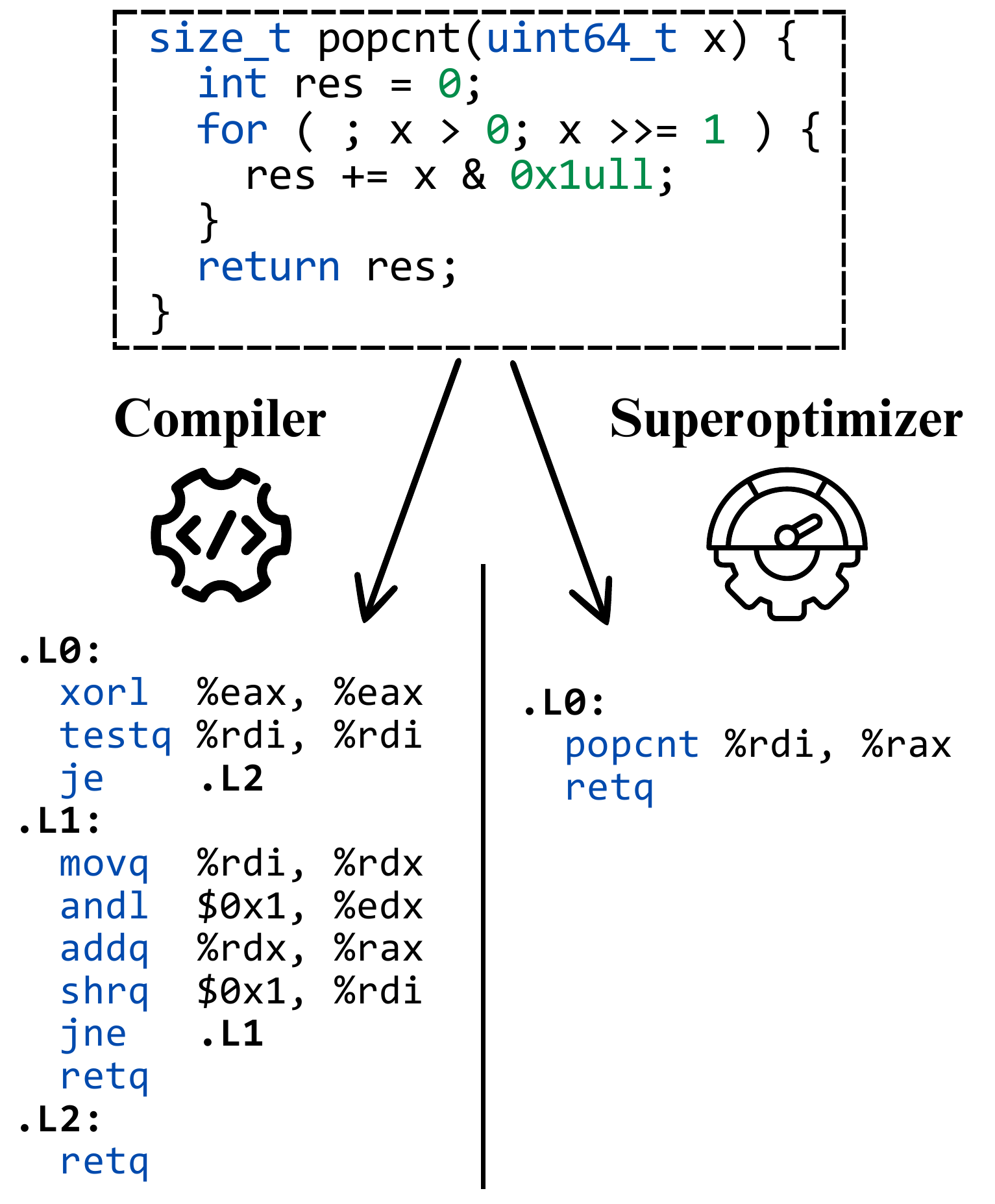}
\caption{\textbf{An equivalent pair from the \ass category in \name.} Both programs are compiled from the same C function shown above, the left using a compiler and the right using a \emph{superoptimizer}. The function counts the number of set bits in the input \CodeIn{\%rdi} register and stores the result in \CodeIn{\%rax}. Their equivalence has been formally verified by the superoptimizer.}
\label{fig:x86}
\end{figure}

\subsection{Pairs from a Superoptimizer (\ass)}
\label{subsec:x86}

\paragraph{Definition of Equivalence.} Two x86-64 assembly programs are considered equivalent if, for any input provided in the specified input registers, both programs produce identical outputs in the specified output registers. Differences in other registers or memory are ignored for equivalence checking.

\paragraph{Example.} \Cref{fig:x86} shows an example of an equivalent program pair in x86-64 assembly. Both programs implement the same C function, which counts the number of bits set to 1 in the variable \CodeIn{x} (mapped to the \CodeIn{\%rdi} register) and stores the result in \CodeIn{\%rax}. The left-hand program, generated by GCC with O3 optimization, uses a loop to count each bit individually, while the right-hand program, produced by a superoptimizer, leverages the \CodeIn{popcnt} instruction, a hardware-supported operation for efficient bit counting. The superoptimizer verifies that both programs are semantically equivalent. Determining this equivalence requires a solid understanding of x86-64 assembly semantics and the ability to reason about all possible bit patterns.

\paragraph{Automation.} A superoptimizer searches a space of programs to find one equivalent to the target. Test cases efficiently prune incorrect candidates, while formal verification guarantees the correctness of the optimized program. Superoptimizers apply aggressive and non-local transformations, making semantic equivalence reasoning more challenging. For example, in \Cref{fig:x86}, while a traditional compiler translates the loop in the source C program into a loop in assembly, a superoptimizer can find a more optimal instruction sequence by leveraging specialized hardware instructions. Such transformations are beyond the scope of traditional compilers.

\paragraph{Dataset Generation.} We use Stoke~\cite{schkufza2013stochastic}  to generate program pairs. Assembly programs are sampled from prior work~\cite{koenig2021adaptive}, and Stoke applies transformations to produce candidate programs. If verification succeeds, the pair is labeled as equivalent; if the generated test cases fail, it is labeled as inequivalent.

\begin{figure}[!tb]
\centering
\includegraphics[width=\columnwidth]{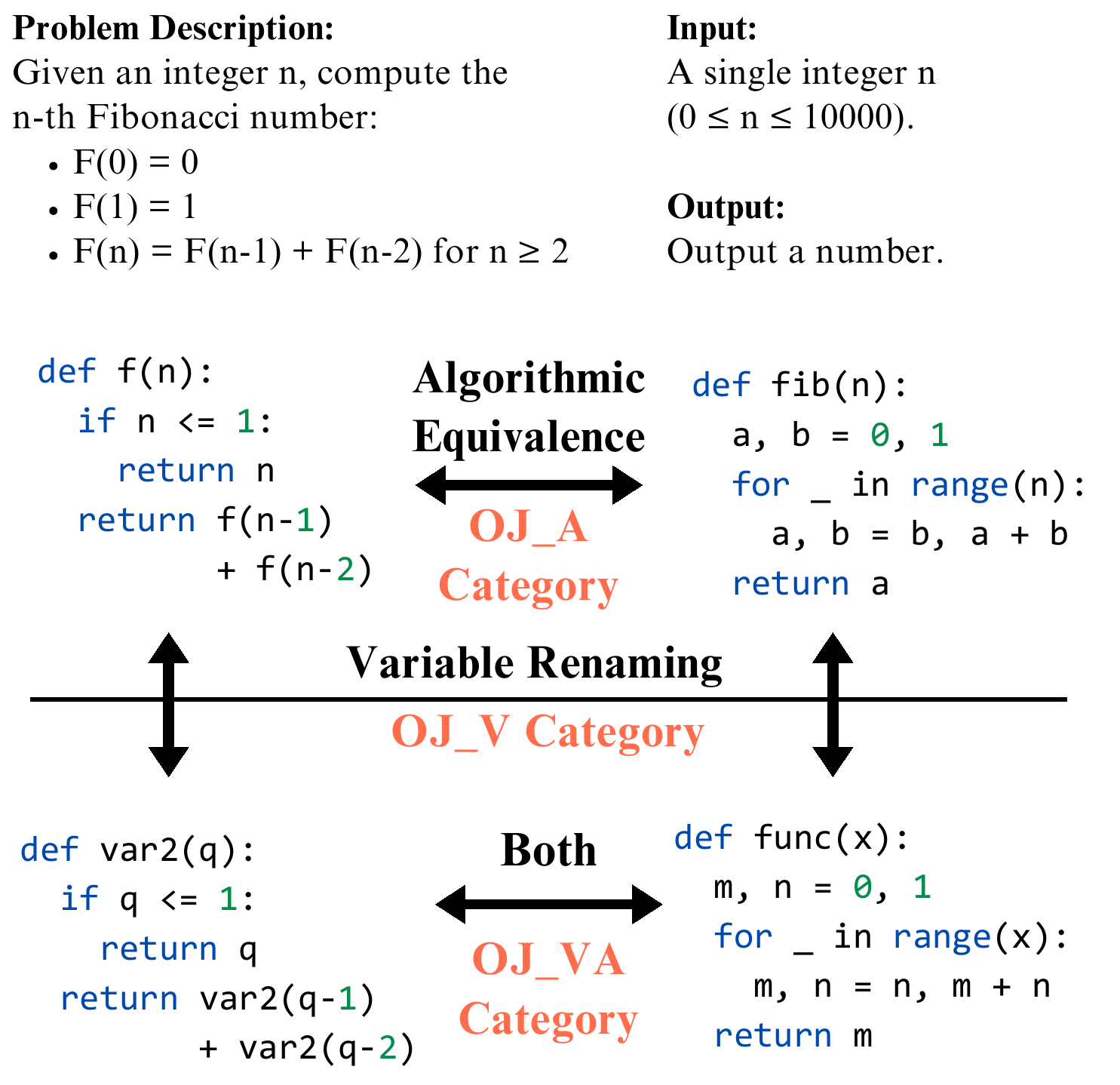}
\caption{\textbf{Equivalent pairs from the \oja, \ojv, \ojva categories in \name.} \oja pairs demonstrate \emph{algorithmic equivalence}, \ojv pairs involve \emph{variable renaming} transformations, and \ojva pairs combine \emph{both} types of variations.}
\label{fig:oj}
\end{figure}

\subsection{Pairs from Programming Contests}
\label{subsec:oj}

\paragraph{Definition of Equivalence.} Two programs are considered equivalent if they solve the same problem by producing the same output for any valid input, as defined by the problem description. Both programs, along with the problem description, are provided to determine equivalence.

\paragraph{Example.} Given the problem description in \Cref{fig:oj}, all four programs are equivalent as they correctly compute the $n$th Fibonacci number. The \textbf{\oja} pairs demonstrate \textbf{algorithmic} equivalence—the left-hand program uses recursion, while the right-hand program employs a for-loop. The \textbf{\ojv} pairs are generated through \textbf{variable renaming}, a purely syntactic transformation that obscures the program’s semantics by removing meaningful variable names. The \textbf{\ojva} pairs combine \textbf{both} algorithmic differences and variable renaming.

\paragraph{Dataset Generation.}
We sample Python submissions using a publicly available dataset from Online Judge (OJ)~\cite{puri2021codenet}. For \oja pairs, accepted submissions are treated as equivalent, while pairs consisting of an accepted submission and a wrong-answer submission are considered inequivalent. Variable renaming transformations are automated with an open-source tool~\cite{PythonTool}.

%% file: 4_setup.tex
\section{Experimental Setup}
\label{sec:experiment}

\paragraph{Dataset.} \name{} consists of 2,400 program pairs across six equivalence categories, each with 200 equivalent and 200 inequivalent pairs. \Cref{tab:dataset} summarizes the statistics of program lengths. Constructing the program pairs required substantial systems effort. For example, for the DCE category, we developed a 2,917-line LLVM-based tool, including 1,472 lines in C and C++, with alias analysis and path feasibility analysis to accurately classify live and dead code.

\begin{table}[!tb]
    \small
    \centering
    \begin{tabular}{l l c c c c}
        \toprule
        \multirow{2}{*}{Category} & \multirow{2}{*}{Language} & \multirow{2}{*}{\# Pairs} & \multicolumn{3}{c}{Lines of Code} \\
        \cmidrule(lr){4-6}
        & & & Min & Max & Avg. \\
        \midrule
        \dce & C & 400 & 98 & 880 & 541 \\
        \cuda & CUDA & 400 & 46 & 1733  & 437  \\
        \ass & x86-64 & 400 & 8  & 29  & 14  \\
        \oja & Python & 400 & 3 & 3403  & 82 \\
        \ojv & Python & 400 & 2 & 4087 &  70 \\
        \ojva & Python & 400 & 3 & 744  & 35 \\
        \bottomrule
    \end{tabular}
    \caption{\textbf{Statistics of the \name{} dataset.}}
    \label{tab:dataset}
\end{table}

\paragraph{Prompts.} The 0-shot evaluation is conducted using the prompt ``You are here to judge if two programs are semantically equivalent. Here equivalence means \{{\em definition}\}. [Program 1]: \{code1\} [Program 2]: \{code2\} Please only output the answer of whether the two programs are equivalent or not. You should only output Yes or No.'' The definition of equivalence and the corresponding program pairs are provided for each category. Additionally, for the categories of \oja, \ojv, and \ojva, the prompt also includes the problem description. The full prompts used in our experiments for each equivalence category are in \Cref{subsec:app:prompt}.

\begin{table*}[!tb]
        \small
        \centering
\begin{tabular}{lccccccc}
\toprule
\textbf{Model} & \textbf{DCE} & \textbf{\cuda} & \textbf{\ass} & \textbf{\oja} & \textbf{\ojv} & \textbf{\ojva} & \textbf{Overall Accuracy} \\
\midrule
\textit{Random Baseline} & \textit{50.0} & \textit{50.0} & \textit{50.0} & \textit{50.0} & \textit{50.0} & \textit{50.0} & \textit{50.0} \\
Llama-3.2-3B-Instruct-Turbo & 50.0 & 49.8 & 50.0 & 51.5 & 51.5 & 51.5 & 50.7 \\
Llama-3.1-8B-Instruct-Turbo & 41.8 & 49.8 & 50.5 & 57.5 & 75.5 & 56.8 & 55.3 \\
Mistral-7B-Instruct-v0.3 & 51.0 & 57.2 & 73.8 & 50.7 & 50.5 & 50.2 & 55.6 \\
Mixtral-8x7B-Instruct-v0.1 & 50.2 & 47.0 & 64.2 & 59.0 & 61.5 & 55.0 & 56.1 \\
Mixtral-8x22B-Instruct-v0.1 & 46.8 & 49.0 & 62.7 & 63.5 & 76.0 & 62.7 & 60.1 \\
Llama-3.1-70B-Instruct-Turbo & 47.5 & 50.0 & 58.5 & 66.2 & 72.0 & 67.5 & 60.3 \\
QwQ-32B-Preview & 48.2 & 50.5 & 62.7 & 65.2 & 71.2 & 64.2 & 60.3 \\
Qwen2.5-7B-Instruct-Turbo & 50.5 & 49.2 & 58.0 & 62.0 & 80.8 & 63.0 & 60.6 \\
gpt-4o-mini-2024-07-18 & 46.8 & 50.2 & 56.8 & 64.5 & 91.2 & 64.0 & 62.2 \\
Qwen2.5-72B-Instruct-Turbo & 42.8 & 56.0 & 64.8 & 72.0 & 76.5 & 70.8 & 63.8 \\
Llama-3.1-405B-Instruct-Turbo & 40.0 & 49.0 & 75.0 & 72.2 & 74.5 & 72.8 & 63.9 \\
DeepSeek-V3 & 41.0 & 50.7 & 69.2 & 73.0 & 83.5 & 72.5 & 65.0 \\
gpt-4o-2024-11-20 & 43.2 & 49.5 & 65.2 & 71.0 & 87.0 & 73.8 & 65.0 \\
claude3.5-sonnet-2024-10-22 & 38.5 & 62.3 & 70.0 & 71.2 & 78.0 & 73.5 & 65.6 \\
claude3.7-sonnet-2025-04-16 & 40.5 & \textbf{63.8} & 64.8 & 70.5 & 89.2 & 73.5 & 67.0 \\
o1-mini-2024-09-12 & 55.8 & 50.7 & 74.2 & 80.0 & 89.8 & 78.8 & 71.5 \\
DeepSeek-R1 & 52.2 & 61.0 & 78.2 & 79.8 & 91.5 & 78.0 & 73.5 \\
o3-mini-2025-01-31 & 68.8 & 59.0 & \textbf{84.5} &  84.2 & 88.2 & 83.2 & 78.0 \\
o4-mini-2025-04-16 & \textbf{76.2} & 60.8 & 83.0 & \textbf{89.0} & \textbf{96.5} & \textbf{88.5} & \textbf{82.3}\\
\midrule
Mean & 49.0 & 53.4 & 66.7 & 68.6 & 78.1 & 68.5 & 64.0 \\
\bottomrule
\end{tabular}
    \caption{\textbf{Accuracy of \numllm models on \name{} under 0-shot prompting.} We report accuracy for each of the six equivalence categories along with the overall accuracy.}
    \label{tab:acc}
\end{table*}

%% file: 5_result.tex
\section{Results}
\label{sec:result}

\subsection{Model Accuracy}
\label{subsec:acc}

\Cref{tab:acc} shows the accuracy results for \numllm state-of-the-art large language models on \name under zero-shot prompting. Our findings are as follows:

\paragraph{Reasoning models achieve the highest performance.} As shown in \Cref{tab:acc}, reasoning models such as OpenAI o3-mini, DeepSeek R1, and o1-mini significantly outperform all others in our evaluation. This further underscores the complexity of equivalence checking, where reasoning models exhibit a distinct advantage.

\paragraph{\name is a challenging benchmark.} Among the \numllm models evaluated, OpenAI o4-mini achieves only \sotalowacc{} in the CUDA category despite being the top-performing model overall, with an accuracy of \sotaacc{}. For the two most difficult categories, the highest accuracy across all models is \sotacuda{} and \sotadce{}, respectively, only modestly above the random baseline of 50\% accuracy for binary classification, highlighting the substantial room for improvement.

\begin{figure}[!tb]
    \centering
    \includegraphics[width=\columnwidth]{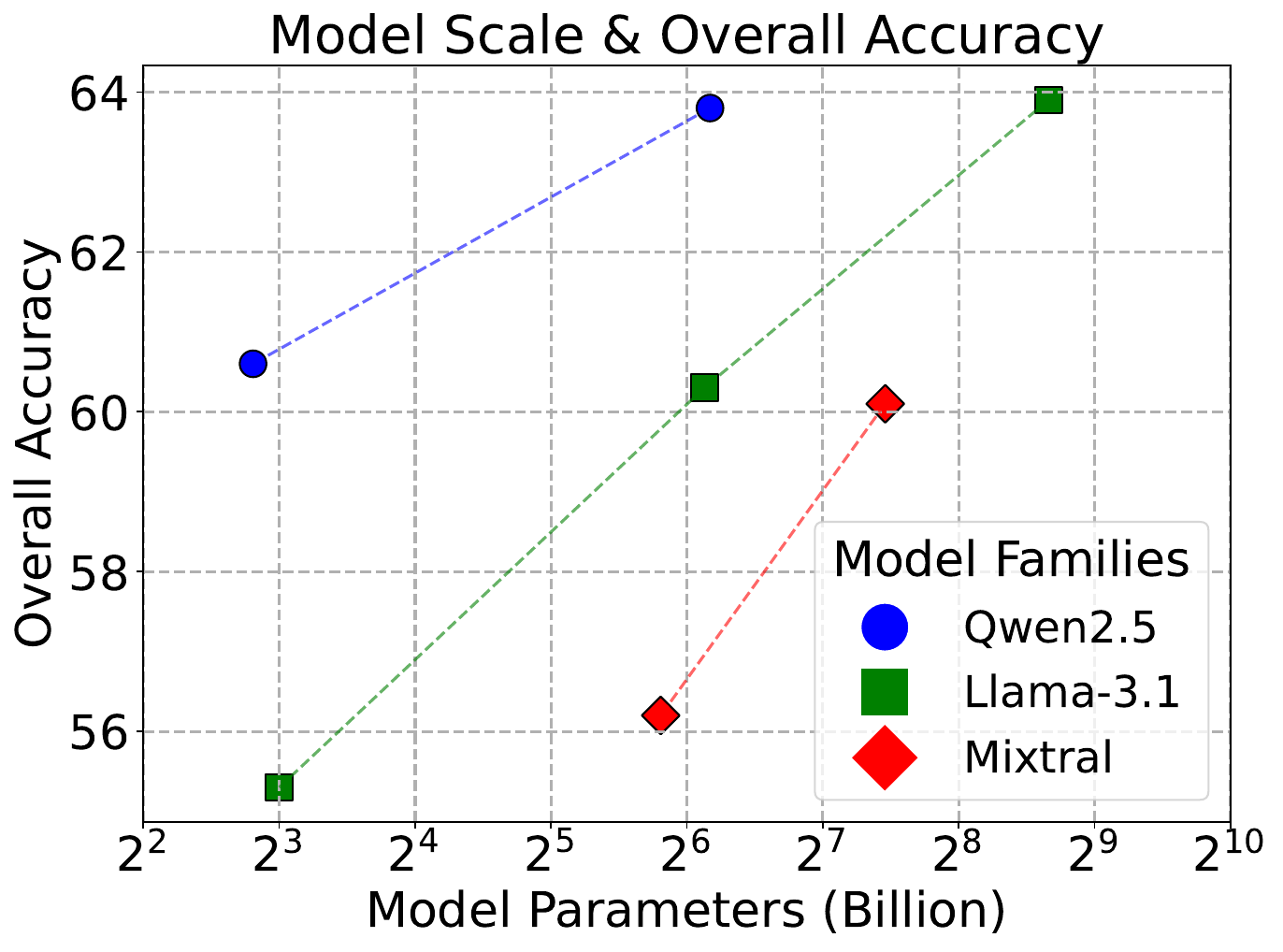}
    \caption{\textbf{Scaling Trend on \name.} Models exhibit consistent gains as parameters increase.}
    \label{fig:scaling}
\end{figure}

\paragraph{Scaling up models improves performance.} Larger models generally achieve better performance. \Cref{fig:scaling} shows scaling trends for the Qwen2.5, Llama-3.1, and Mixtral families, where accuracy improves with model size. The x-axis is on a logarithmic scale, highlighting how models exhibit consistent gains as parameters increase.

\subsection{Difficulty Analysis}
\label{subsec:difficulty}

We conduct a detailed difficulty analysis across equivalence categories and study how syntactic similarity influences model predictions.

\paragraph{Difficulty by Transformation Type.} Each category adopts a specific definition of equivalence (see \Cref{sec:method}), and the program transformations used in each category differ accordingly. We find that purely syntactic transformations are substantially easier for models, while structural and compiler-involved transformations are much harder. Specifically, \textbf{OJ\_V} (variable renaming) achieves the highest mean accuracy of 78.1\%, as it only requires surface-level reasoning. \textbf{OJ\_A} (algorithmic equivalence) and \textbf{OJ\_VA} (variable renaming combined with algorithmic differences) achieve similar accuracies of 68.6\% and 68.5\%, respectively. In contrast, \textbf{x86-64} (66.7\%) and \textbf{CUDA} (53.4\%) involve complex instruction-level or memory-level transformations, requiring deeper semantic reasoning. \textbf{DCE} (dead code elimination) is the most difficult category, with a mean accuracy of 49.0\%, suggesting that models struggle with nuanced program analysis concepts. 

\paragraph{Difficulty by Syntactic Similarity.}
To assess whether LLM predictions reflect understanding of program semantics rather than reliance on surface-level syntax, we analyze how syntactic similarity affects model behavior. Using Moss~\citep{schleimer2003winnowing}, a plagiarism detection tool, we observe the following:

\begin{itemize}
    \item For program pairs with low syntactic similarity, models tend to predict ``inequivalent,'' even when the programs are semantically equivalent. This suggests an overreliance on the superficial form of the code.
    \item For syntactically similar pairs, models are more likely to predict ``equivalent,'' indicating a tendency to associate similarity in form with equivalence in program semantics.
\end{itemize}

We validate this trend through statistical testing: at significance level~\((\alpha = 0.05)\), model accuracy on equivalent pairs increases with syntactic similarity, while accuracy on inequivalent pairs decreases. This disconnect between syntactic form and execution behavior, as discussed in \Cref{sec:intro}, suggests that models do not fully grasp program semantics.

\paragraph{Implications for Benchmark Design.}
These findings suggest that future benchmarks should emphasize \emph{syntactically dissimilar yet equivalent program pairs} and \emph{syntactically similar yet inequivalent program pairs} to create more challenging and diagnostic benchmarks for evaluating the deep semantic reasoning capabilities of LLMs.

\subsection{Bias in Model Prediction}
\label{subsec:bias}

We evaluate the prediction bias of the models and observe a pronounced tendency to misclassify equivalent programs as inequivalent in the \cuda and \ass categories. \Cref{tab:bias} presents the results for four representative models, showing high accuracy for inequivalent pairs but significantly lower accuracy for equivalent pairs, with full results for all models in \Cref{subsec:app:bias}.

The bias in the \cuda category arises from extensive structural transformations, such as loop restructuring and shared memory optimizations, which make paired programs appear substantially different. In the \ass category, superoptimization applies non-local transformations to achieve optimal instruction sequences, introducing aggressive code restructuring that complicates equivalence reasoning and leads models to misclassify equivalent pairs as inequivalent frequently.

\begin{table}[!h]
    \centering
\small
\begin{tabular}{lcccc}
\toprule
\multicolumn{1}{c}{\multirow{2}{*}{\textbf{Model}}} & \multicolumn{2}{c}{\textbf{\cuda}} & \multicolumn{2}{c}{\textbf{\ass}} \\
\cmidrule(lr){2-5}
 & \textbf{Eq} & \textbf{Ineq} & \textbf{Eq} & \textbf{Ineq} \\
\midrule
\textit{Random Baseline} & \textit{50.0} & \textit{50.0} & \textit{50.0} & \textit{50.0} \\
o3-mini & 27.5 & 90.5 & 69.5 & 99.5 \\
o1-mini & 2.5 & 99.0 & 50.0 & 98.5 \\
DeepSeek-R1 & 28.0 & 94.0 & 57.5 & 99.0 \\
DeepSeek-V3 & 8.5 & 93.0 & 44.0 & 94.5 \\
\bottomrule
\end{tabular}
    \caption{Accuracies on equivalent and inequivalent pairs in the \cuda and \ass categories under 0-shot prompting, showing that \textbf{models perform significantly better on inequivalent pairs}. Random guessing serves as an unbiased baseline for comparison. More results are in \Cref{subsec:app:bias}.}
\label{tab:bias}
\end{table}

\subsection{Prompting Strategies Analysis}
\label{subsec:prompt}

We study few-shot in-context learning and Chain-of-Thought (CoT) prompting, evaluating four strategies: 0-shot, 4-shot, 0-shot with CoT, and 4-shot with CoT. For 4-shot, prompts include 2 equivalent and 2 inequivalent pairs. \Cref{tab:prompt} shows the results.

Our key finding is that prompting strategies barely improve performance on \name, highlighting the difficulty of understanding program semantics.

\begin{table}[!tb]
\centering
\small
\begin{tabular}{lcccc}
\toprule
\textbf{Model} & \textbf{0S} & \textbf{4S} & \textbf{0S-CoT} & \textbf{4S-CoT} \\
\midrule
o1-mini & 71.5 & 71.5 & \textbf{71.9} & \textbf{71.9} \\
gpt-4o & 65.0 & \textbf{66.5} & 62.5 & 62.7 \\
DeepSeek-V3 & 65.0 & \textbf{66.9} & 63.3 & 62.5 \\
gpt-4o-mini & 62.2 & \textbf{63.5} & 60.2 & 61.2 \\
\bottomrule
\end{tabular}
\caption{\textbf{Accuracies of different prompting techniques.} We evaluate 0-shot and 4-shot in-context learning, both without and with Chain-of-Thought (CoT). Prompting strategies barely improve performance.}
\label{tab:prompt}
\end{table}

%% file: 6_discussion.tex
\section{Discussion and Future Directions}
\label{sec:discussion}

\paragraph{Scope and Positioning}
Machine learning has been applied to many code-related tasks, such as clone detection~\cite{white2016deep}, code search~\cite{gao2024virtual}, and bug finding~\cite{deng2023large}. \name{} focuses on equivalence checking, which differs fundamentally by evaluating a model's understanding of program semantics. Unlike natural language, code is executable, and its correctness depends on execution results rather than form. For example, clone detection captures syntactic or structural similarity without considering behavior. In contrast, \name{} tests whether two programs produce the same outputs for all inputs, offering an informative benchmark for reasoning about program behavior.

\paragraph{Developer Use Cases}
\name{} evaluates whether LLMs truly understand program semantics, a capability that underpins downstream tasks such as program optimization, software refactoring, and transpilation. These tasks are central to practical scenarios where coding assistants must propose improvements or transformations without changing program behavior. For example, after a developer performs a refactoring, a coding assistant that performs well on \name{} would be better positioned to judge whether the transformed code preserves the same functionality as the original.

\paragraph{Labeling Soundness}
To ensure high-assurance equivalence labels, \name{} relies on transformations grounded in program analysis, compiler scheduling, and superoptimization, all of which offer strong soundness guarantees. In contrast, approaches such as random testing~\cite{jiang2009automatic}, similarity-based tools~\cite{silva2017refdiff}, and refactoring datasets lack formal guarantees and risk introducing incorrect labels.

\paragraph{Design and Extensibility}
\name{} is designed with modularity in mind: each equivalence category corresponds to a distinct class of program transformations. Demonstrating strong performance in these settings would indicate that LLMs could support some components of compiler pipelines (e.g., dead code elimination (DCE) as a core compiler optimization, or CUDA program scheduling for high-performance ML systems). We focus on the six categories where large-scale, high-confidence labels can be generated automatically. That said, equivalence checking is a general task that is applicable to all programming languages. We view our benchmark as a first step, and its modular design allows future extension to additional categories and languages.

\paragraph{Evaluation of Reasoning Trace}
While our evaluation centers on binary classification, understanding the rationale behind model predictions is an important direction. Explanations may take the form of natural language or formal proofs, but verifying their correctness remains difficult. Natural language lacks reliable automated validation, since using LLMs as judges can produce unsound results. Building a proof-based evaluation framework using tools such as Lean is also highly nontrivial. We present a manual case analysis of reasoning trace correctness in \Cref{subsec:case} and leave automated robust evaluation of reasoning as future work.

\paragraph{Effect of Fine-Tuning}
We tested whether supervised fine-tuning improves performance. Fine-tuning Qwen2.5-14B-Instruct with LoRA for 3 epochs on 1,200 labeled examples increased accuracy from 59.8\% to 63.2\%. The small gain suggests that binary labels alone provide limited learning signals for reasoning about program semantics. Prior work has explored training with \emph{program execution traces} to better capture execution behavior. We conducted an additional experiment to evaluate the training approach from SemCoder~\cite{ding2024semcoder}. The base model (DeepSeek-Coder-6.7B) achieves 49.9\% accuracy on our benchmark, and the fine-tuned model released by SemCoder reaches 54.9\%. While this shows some benefit, the improvement remains modest. These results support our broader claim: \name{} presents a difficult and meaningful challenge even for fine-tuned models, and deeper semantic understanding remains out of reach for current approaches.

\paragraph{Future Directions}
We believe \name{} can inform future research on task-specific training methods, including: (1) distilling reasoning traces from stronger models, (2) scaling training with larger datasets generated through our pipeline, (3) developing agentic approaches where LLMs actively execute and compare programs using tools (e.g., a Python interpreter) to generate inputs that expose differences, (4) applying reinforcement learning with execution-based feedback, and (5) creating datasets with program analysis concepts (see \Cref{subsec:case}) for training LLMs.

%% file: 7_conclusion.tex
\section{Conclusion}

\name{} is a benchmark for evaluating whether large language models (LLMs) truly understand program semantics. We propose the task of equivalence checking, which asks whether two programs produce identical outputs for all possible inputs, as a direct way to test a model’s ability to reason about program behavior. The dataset consists of 2400 program pairs across four languages and six categories, constructed through a fully automated pipeline that provides high-confidence labels and nontrivial difficulty. Our evaluation of \numllm state-of-the-art LLMs shows that even the best-performing models achieve only modest accuracy in the most challenging categories. Further analysis shows that LLMs often rely on syntactic similarity instead of demonstrating robust reasoning about program semantics, underscoring the need for further advances in the semantic understanding of programs.

%% file: 8_appendix.tex
\section{Appendix}
\label{sec:appendix}

\subsection{Case Studies}
\label{subsec:case}

\paragraph{Models lack capabilities for sound equivalence checking.} We find that simple changes that lead to semantic differences can confuse the models, causing them to produce incorrect predictions despite their correct predictions on the original program pairs. For example, o3-mini, which is one of the top-performing models in \cuda category, can correctly classify the pair shown in \Cref{fig:cuda} as equivalent. Next, we introduce synchronization bugs into the right-hand program, creating two inequivalent pairs with the original left-hand program: (1) removing the first \CodeIn{\_\_syncthreads();} allows reads before all writes complete, causing race conditions; (2) removing the second \CodeIn{\_\_syncthreads();} lets faster threads overwrite shared data while slower threads read it. Despite these semantic differences, o3-mini misclassifies both pairs as equivalent.

\paragraph{Proper hints enable models to correct misjudgments.} After o3-mini misclassifies the modified pairs, a hint about removed synchronization primitives allows it to correctly identify both as inequivalent, with accurate explanations highlighting data races. This suggests that training models on dedicated program analysis datasets, beyond only raw source code, may be useful for improving their code reasoning capabilities.

\subsection{Model Prediction Bias}
\label{subsec:app:bias}

We evaluate the prediction bias of the models and observe a pronounced tendency to misclassify equivalent programs as inequivalent in the \cuda and \ass categories. \Cref{fig:appbias} here shows the full results on models under 0-shot prompting.

\begin{figure*}[!tb]
\centering
\small
\begin{tabular}{lcccc}
\toprule
\multicolumn{1}{c}{\multirow{2}{*}{\textbf{Model}}} & \multicolumn{2}{c}{\textbf{\cuda}} & \multicolumn{2}{c}
{\textbf{\ass}} \\
\cmidrule(lr){2-5}
 & \textbf{Eq} & \textbf{Ineq} & \textbf{Eq} & \textbf{Ineq} \\
\midrule
\textit{Random Baseline} & \textit{50.0} & \textit{50.0} & \textit{50.0} & \textit{50.0} \\
deepseek-ai/DeepSeek-V3 & 8.5 & 93.0 & 44.0 & 94.5 \\
deepseek-ai/DeepSeek-R1 & 28.0 & 94.0 & 57.5 & 99.0 \\
meta-llama/Llama-3.1-405B-Instruct-Turbo & 6.0 & 92.0 & 68.5 & 81.5 \\
meta-llama/Llama-3.1-8B-Instruct-Turbo & 2.0 & 97.5 & 1.0 & 100.0 \\
meta-llama/Llama-3.1-70B-Instruct-Turbo & 7.0 & 93.0 & 27.5 & 89.5 \\
meta-llama/Llama-3.2-3B-Instruct-Turbo & 0.0 & 99.5 & 0.0 & 100.0 \\
anthropic/claude-3-5-sonnet-20241022 & 62.5 & 62.0 & 49.5 & 90.5 \\
Qwen/Qwen2.5-7B-Instruct-Turbo & 18.5 & 80.0 & 17.5 & 98.5 \\
Qwen/Qwen2.5-72B-Instruct-Turbo & 14.5 & 97.5 & 36.0 & 93.5 \\
Qwen/QwQ-32B-Preview & 35.0 & 66.0 & 39.0 & 86.5 \\
mistralai/Mixtral-8x7B-Instruct-v0.1 & 18.0 & 76.0 & 50.5 & 78.0 \\
mistralai/Mixtral-8x22B-Instruct-v0.1 & 10.5 & 87.5 & 32.5 & 93.0 \\
mistralai/Mistral-7B-Instruct-v0.3 & 52.5 & 62.0 & 87.0 & 60.5 \\
openai/gpt-4o-mini-2024-07-18 & 0.5 & 100.0 & 16.5 & 97.0 \\
openai/gpt-4o-2024-11-20 & 0.0 & 99.0 & 68.5 & 62.0 \\
openai/o3-mini-2025-01-31 & 27.5 & 90.5 & 69.5 & 99.5 \\
openai/o1-mini-2024-09-12 & 2.5 & 99.0 & 50.0 & 98.5 \\
\bottomrule
\end{tabular}
\caption{Model prediction bias.}
\label{fig:appbias}
\end{figure*}

\subsection{Prompts}
\label{subsec:app:prompt}

\lstset{style=mystyle}

\subsubsection{DCE Category}
We show the prompts for the 0-shot setting.

You are here to judge if two C programs are semantically equivalent.\\
    Here equivalence means that, when run on the same input, the two programs always have the same program state at all corresponding points reachable by program execution.\\
    \text{    [Program 1]}:\\
    \begin{lstlisting}
    {program_1_code}
    \end{lstlisting}
    \text{    [Program 2]}:\\
    \begin{lstlisting}
    {program_2_code}
    \end{lstlisting}
    
    Please only output the answer of whether the two programs are equivalent or not. You should only output YES or NO.\\
\\

\subsubsection{CUDA Category}
We show the prompts for the 0-shot setting.

You are here to judge if two CUDA programs are semantically equivalent.\\
Here equivalence means that, when run on the same valid input, the two programs always compute the same mathematical output (neglecting floating point rounding errors).\\
\text{    [Program 1]}:
\begin{lstlisting}
{program_1_code}
\end{lstlisting}
\text{    [Program 2]}:
\begin{lstlisting}
{program_2_code}
\end{lstlisting}

Please only output the answer of whether the two programs are equivalent or not. You should only output YES or NO.\\

\subsubsection{x86-64 Category}
We show the prompts for the 0-shot setting.

You are here to judge if two x86-64 programs are semantically equivalent.\\
    Here equivalence means that, given any input bits in the register \texttt{\{def\_in\}}, the two programs always have the same bits in register \texttt{\{live\_out\}}. Differences in other registers do not matter for equivalence checking.\\

\text{[Program 1]:}\\
\begin{lstlisting}
{program_1_code}
\end{lstlisting}

\text{[Program 2]:}\\
\begin{lstlisting}
{program_2_code}
\end{lstlisting}

Please only output the answer of whether the two programs are equivalent or not. You should only output YES or NO.\\

\subsubsection{OJ\_A, OJ\_V, OJ\_VA Category}
We show the prompts for the 0-shot setting.

You are here to judge if two Python programs are semantically equivalent.\\
    You will be given \text{[Problem Description]}, \text{[Program 1]} and \text{[Program 2]}.\\
    Here equivalence means that, given any valid input under the problem description, the two programs will always give the same output.\\

\noindent\text{[Problem Description]}:\\
\begin{lstlisting}
{problem_html}
\end{lstlisting}

\noindent\text{[Program 1]}:\\
\begin{lstlisting}
{program_1_code}
\end{lstlisting}

\noindent\text{[Program 2]}:\\
\begin{lstlisting}
{program_2_code}
\end{lstlisting}

Please only output the answer of whether the two programs are equivalent or not. You should only output YES or NO.